\documentclass[10pt,twocolumn,letterpaper]{article}

\usepackage{wacv}              

\usepackage{graphicx}
\usepackage{amsmath}
\usepackage{amssymb}
\usepackage{booktabs}
\usepackage{multirow}

\usepackage[pagebackref,breaklinks,colorlinks]{hyperref}

\usepackage[capitalize]{cleveref}
\crefname{section}{Sec.}{Secs.}
\Crefname{section}{Section}{Sections}
\Crefname{table}{Table}{Tables}
\crefname{table}{Tab.}{Tabs.}

\def\wacvPaperID{407} 
\def\confName{WACV}
\def\confYear{2024}

\begin{document}

\title{Generalizing to Unseen Domains in Diabetic Retinopathy Classification}

\author{Chamuditha Jayanga Galappathththige\hspace{8pt}  Gayal Kuruppu\hspace{8pt}  Muhammad Haris Khan\\
Mohamed bin Zayed University of Artificial Intelligence, UAE.\\
{\tt\small \{chamuditha.jayanga,gayal.kuruppu,muhammad.haris\}@mbzuai.ac.ae}
}
\maketitle

\begin{abstract}
   Diabetic retinopathy (DR) is caused by long-standing diabetes and is among the fifth leading cause for visual impairment. The prospects of early diagnosis and treatment could be helpful in curing the disease, however, the detection procedure is rather challenging and mostly tedious. Therefore, automated diabetic retinopathy classification using deep learning techniques has gained interest in the medical imaging community. Akin to several other real-world applications of deep learning, the typical assumption of i.i.d data is also violated in DR classification that relies on deep learning. Therefore, developing DR classification methods robust to unseen distributions is of great value. In this paper, we study the problem of generalizing a model to unseen distributions or domains (a.k.a domain generalization) in DR classification. To this end, we propose a simple and effective domain generalization (DG) approach that achieves self-distillation in vision transformers (ViT) via a novel prediction softening mechanism. This prediction softening is an adaptive convex combination of one-hot labels with the model’s own knowledge. We perform extensive experiments on challenging open-source DR classification datasets under both multi-source and more challenging single-source DG settings with three different ViT backbones to establish the efficacy and applicability of our approach against competing methods. For the first time, we report the performance of several state-of-the-art domain generalization (DG) methods on open-source DR classification datasets after conducting thorough experiments. Finally, our method is also capable of delivering improved calibration performance than other methods, showing its suitability for safety-critical applications, including healthcare. We hope that our contributions would instigate more DG research across the medical imaging community. Code is available at \href{https://github.com/Chumsy0725/SPSD-ViT}{github.com/Chumsy0725/SPSD-ViT}.

\end{abstract}


\begin{figure*}[!htp]
\begin{center}
\includegraphics[scale=0.62]{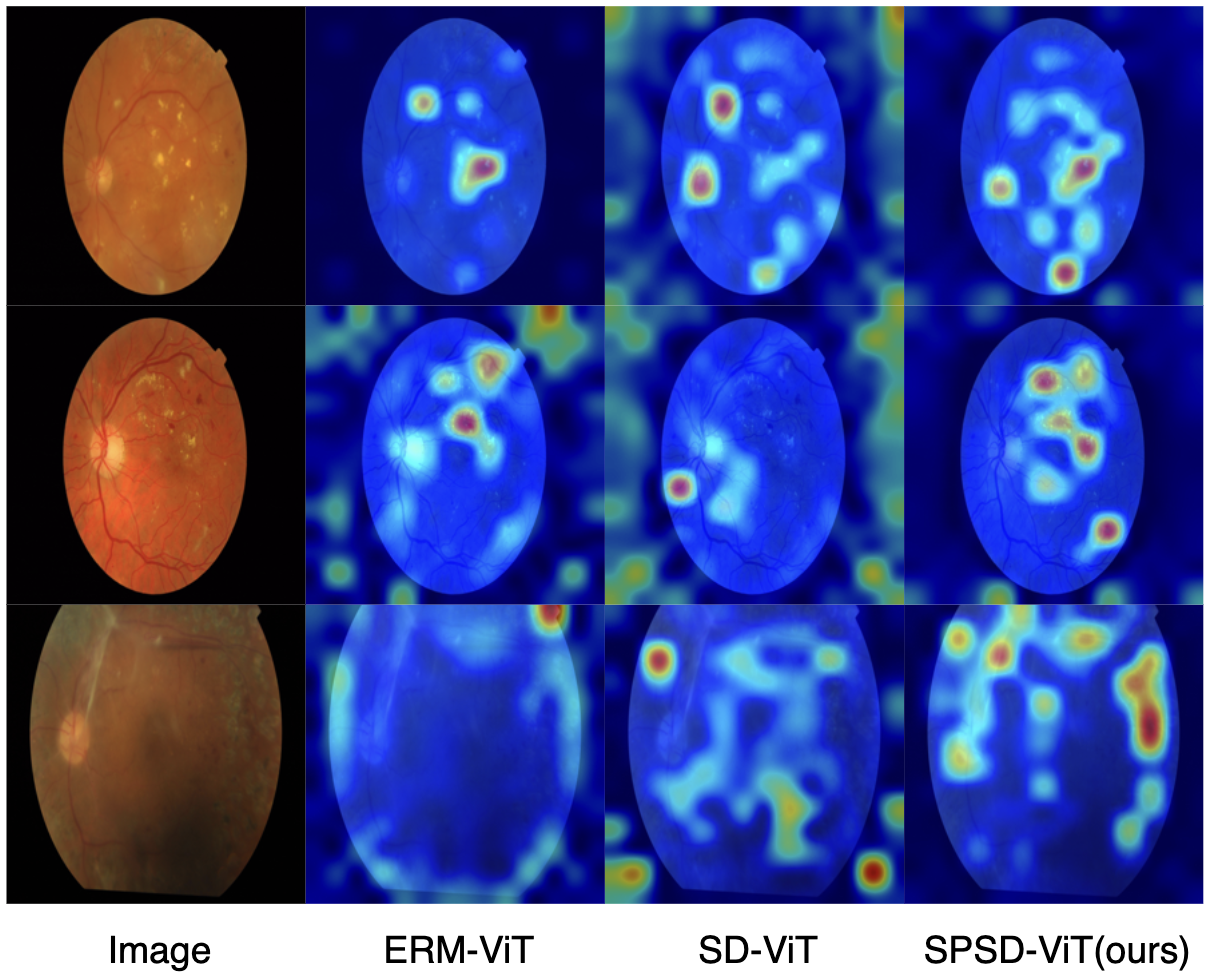}
\end{center}
   \caption{Attention
heatmaps obtained from the final ViT block. We can observe that the baselines: ERM-ViT and SD-ViT are suspectible to relying on non-generalizable domain-specific features such as the background. On the other hand, our method (SPSD-ViT) is capable of exploiting cross-domain generalizable features which mostly correspond to haemorrhages, microaneurysms, exudates and cotton wool spots.}



\label{fig:motivation_fig}
\end{figure*}

\section{Introduction}
\label{sec:intro}

Diabetic retinopathy (DR) is a global medical problem over the last few decades; it arises from a complication in Diabetes Mellitus. In DR, the development of glucose in blood vessels can block them, thereby causing possible swelling and/or leaking of blood or fluid which could eventually lead to visual impairment \cite{nair2019categorization,nair2019classification}. Approximately 33\% of 285 million people with diabetes mellitus across the globe have signs of DR \cite{kumar2016automated}. 
The prevailing practice requires doctors to manually examine the fundus images of the eye to understand the severity of DR. However, this is a time-consuming process and also there is a scarcity of medical professionals compared to the number of patients. Therefore, without much surprise, the development of AI-powered tools capable of accurately detecting DR has gained importance in the recent past.
Several studies in this pursuit utilize fundus images, which visually depict the current ophthalmic appearance of a person’s retina \cite{sebastian2023survey}. The existence of DR symptoms in these fundus images can be used to classify them using several steps such as retinal blood vessel segmentation, lesion segmentation, and DR detection \cite{raja2019automatic}. We can detect DR and its current stage by examining the presence/absence of several lesions. The lesions that are important for diagnosis are microaneurysms (MAs), superficial retinal hemorrhages (SRHs), exudates (Exs) both soft exudates (SEs) and hard exudates (HEs) intraretinal hemorrhages (IHEs), and cotton wool spots (CWSs) \cite{sebastian2023survey}. According to recent studies, DR can be classified into five different categories: namely no DR, mild DR, moderate DR, severe DR, and proliferative DR \cite{kempen2004prevalence}.

The typical assumption of i.i.d data which belong to training and testing sets is often violated in many real-world applications of deep learning, e.g., methods for DR classification \cite{atwany2022drgen}. Unsupervised domain adaptation is a line of research for handling domain shift \cite{ganin2015unsupervised,sun2016deep,long2016unsupervised,kang2019contrastive,yan2020improve}, but it requires the availability of unlabelled data and an adaptation phase, which typically consists of model re-training. Such requirements are often difficult to fulfill in most medical diagnosis applications. 
A viable direction is domain generalization (DG), which does not require the availability of target domain data and any adaptation phase and therefore it is more rewarding but also challenging \cite{motiian2017unified,li2017deeper,carlucci2019domain,li2019episodic,zhou2020learning,khan2021mode}. Unfortunately, very little attention is paid to the development of DG methods for DR classification \cite{atwany2022drgen}, which will play a pivotal role in realizing robust DR classification systems. To this end, we explore the problem of generalizing to unseen domains (DG) for the DR classification tasks. We summarise our key contributions as follows:

\noindent\textbf{Contributions:} \textbf{(1)} Owing to the increasing popularity of vision transformers (ViTs) \cite{dosovitskiy2020image,touvron2021training,tuli2021convolutional,wu2021cvt}, to our knowledge, we propose a first ViT-based DG approach for DR classification that self-distills the knowledge of the full network to its intermediate blocks via a new prediction softening mechanism. \textbf{(2)} We construct softened predictions by formulating an adaptive convex combination of one-hot labels with the model's own knowledge. It strengths intermediate representations and induces model regularization to alleviate the overfitting to source domains, thereby encouraging the learning of more robust (typically domain-invariant) features and reduces reliance on non-robust (typically domain-specific) features (Fig.~\ref{fig:motivation_fig}).
\textbf{(3)} We conduct experiments on challenging DR datasets following Domainbed \cite{Gulrajani2021InSO} protocol under both multi-source and single-source DG settings with three different ViT backbones. Results show the efficacy and applicability of our approach against baselines and established methods. 
\textbf{(4)} For the first time, we report the performance of several state-of-the-art DG methods on open-source DR classification datasets.

\section{Related Work}
\label{section:Related Work}
\noindent\textbf{Domain Generalization:} 
Domain generalization \cite{li2017deeper,carlucci2019domain,li2019episodic,khan2021mode,Lv_2022_CVPR, Wang_2023_CVPR}  utilizes data from multiple source domains for training to generalize to a new (unseen) domain. Some DG methods explicitly aimed at reducing domain gap in the feature space \cite{muandet2013domain,ganin2016domain,li2018domain, ding2022domain, Zhang_2022_CVPR}. Another class of methods attempted the learning of generalizable model parameters through variants of meta-learning \cite{Li2018MLDG,dou2019domain,balaji2018metareg,li2019episodic, Chen_2022_CVPR, Qin_2023_CVPR}. Through leveraging different auxiliary tasks, some methods were proposed to robustify the model against domain shifts \cite{carlucci2019domain,wang2020learning}. Furthermore, various DG methods resorted to devising data augmentation techniques for improving cross-domain generalization \cite{shankar2018generalizing,zhou2020learning,khan2021mode, Kang_2022_CVPR} while some employ test time adaptation methods\cite{Chen_2023_CVPR, 10.1007/978-3-031-25085-9_17}. Recently, Gulrajani et al. \cite{Gulrajani2021InSO} proposed a new benchmark for DG, named ``Domainbed'', which includes a rigorous evaluation protocol that ensures a fair comparison between different DG algorithms. It showed that even a simple Empirical Risk Minimization (ERM) method can be competitive with many of the current state-of-the-art DG approaches. As such, the Domainbed protocol has quickly gained popularity and is now considered a standard for evaluating DG algorithms. In this study, we have also adopted the Domainbed protocol to report results. By using this protocol, we can provide a reliable and fair comparison of our DG approach with other state-of-the-art methods. Our results demonstrate the effectiveness of our approach, even under the strict evaluation criteria of the Domainbed benchmark.

A few methods have proposed variants of teacher-student formulation \cite{hinton2015distilling} to tackle the DG problem. \cite{wang2021embracing} proposed a CNN-based teacher-student distillation scheme along with  a gradient filter as an efficient regularization term. Recently \cite{sultana2022self} developed a self-distillation strategy to improve the DG capabilities of ViTs. It scales the logits of both the full network and a randomly sampled block with a fixed temperature parameter prior to distillation. We adapt this self-distillation for ViTs in DG for DR classification and propose a new prediction softening mechanism, featuring an adaptive convex combination of zero-entropy labels with the model's own knowledge. We empirically show that it is more effective in improving the model's generalizability to unseen domains in DR classification.

\noindent\textbf{DG in medical image analysis:} 
The distribution of data originating from different hospitals or even different sensors could be sparingly different and hence it is important that a model should generalize to a different data distribution than it is trained on. Despite carrying significant importance, DG for medical imaging analysis remains largely unexplored. Among a few DG methods, \cite{li2022domain} developed a meta-leaning approach based on episodic training with task augmentation for medical image classification, and \cite{li2020domain} leveraged variational encoding to realize a characteristic feature space through linear-dependency regularization. 
DG in medical imaging has also been explored in the context of Federated Learning (FL). 
To allow privacy-protected distribution of information among clients, \cite{liu2021feddg} presented episodic learning in Continuous Frequency Space (ELCFS) approach. We note that there is very little work on studying domain generalization for DR classification. Recently, \cite{atwany2022drgen} proposed the very first approach for robustifying the model under data from unseen domains in DR classification. It achieves flatness during the training of convolutional neural network (CNN) and also employs domain-level gradient variance regularization. We also propose a new DG approach for DR categorization which transfers the model's (ViTs) full knowledge to its intermediate feature routes by a new prediction softening scheme.

\section{Proposed Method}
\label{section:Proposed Method}
\subsection{Preliminaries}
\label{subsection: Preliminaries}

\noindent\textbf{DG problem settings:}
In the typical domain generalization (DG) setting, as outlined in \cite{Gulrajani2021InSO}, we assume access to data from a set of training (source) domains, denoted as $\mathcal{D}=\{\mathcal{D}\}_{n=1}^{N}$. 
Each domain $\mathcal{D}_{n}$ represents a distribution over the input space $\mathcal{X}$, and there are a total of $N$ training domains. 
From each domain $\mathcal{D}_{n}$, we sample $K$ training images consisting of pairs of inputs $\mathbf{x}_{n}^{k} \in \mathcal{X}$ and labels $y_{n}^{k}\in \mathcal{Y}$, where $k$ ranges from 1 to $K$. Moreover, we assume the existence of a set of target domains, denoted as $\{\mathcal{T}\}_{t=1}^{T}$, where $T$ is the total number of target domains and is typically 1. The core objective in DG is to learn a function ${\mathcal{F}}_{\theta}: \mathcal{X} \rightarrow \mathcal{Y}$, parameterized by $\theta$ which is capable of predicting accurate labels for input data from an unseen target domain $\mathcal{T}_{t}$.

\begin{figure*}[!htp]
\begin{center}
\includegraphics[scale=0.42]{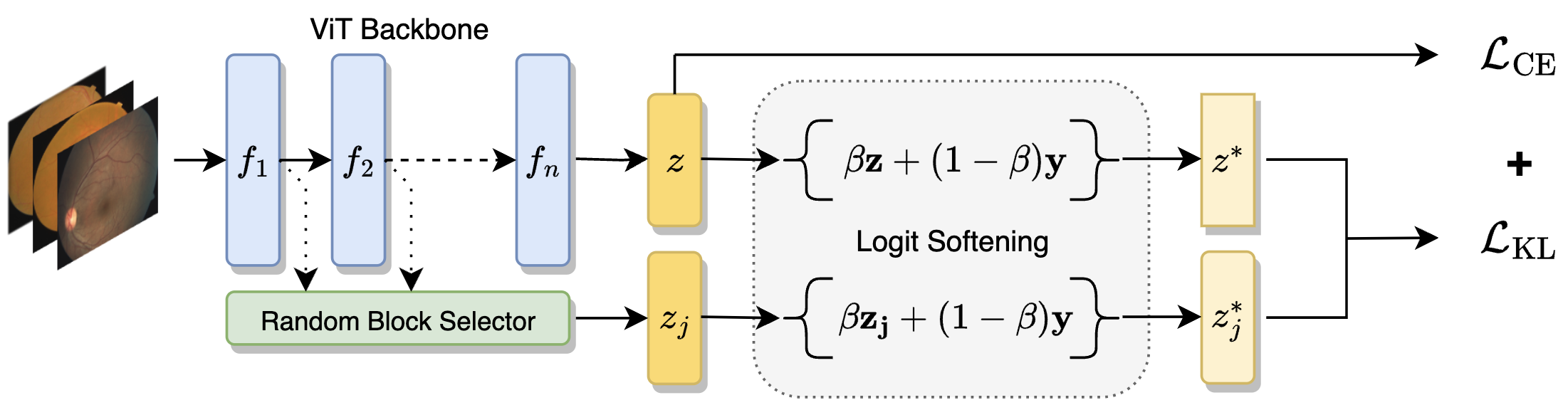}
\end{center}
   \caption{Overall architecture of our proposed self-distilled ViT with the prediction softening mechanism.}
\label{fig:overall_architecture}
\end{figure*}

\noindent\textbf{ViT-based ERM for DG:} We first briefly revisit empirical risk minimization (ERM) in the context of DG and then describe the ViT-based ERM in DG for DR classification task. We assume the availability of a loss function $\mathcal{L}$ that can measure the discrepancy between the predicted label and the desired label. The ERM for DG accumulates data from all training (source) domains and trains a classifier that finds a predictor by minimizing \cite{vapnik1999nature}: $ \frac{1}{M}\sum_{i=1}^{M}\mathcal{L}(\mathcal{F}_{\theta}(\mathbf{x}^{i},y^{i}))$. 
Where $M=N \times K$ denotes the total number of images from all training (source) domains. The work of \cite{Gulrajani2021InSO} established that this simple ERM-based DG baseline reveals competitive performance against many preceding state-of-the-art DG methods under a fair evaluation protocol. 


Now we assume that the model $\mathcal{F}_{\theta}$ in ERM-DG (for DR classification) contains $J$ intermediate blocks/layers and a final classifier $h$, which can be written as: $\mathcal{F} = (f_{1} \circ f_{2} \circ f_{3} \circ \ldots f_{J}) \circ h$, where $f_{j}$ denotes an intermediate block/layer. If this model is a ViT (e.g., ~DeIT-Small \cite{touvron2021training}), then $f_{j}$ is a self-attention transformer block. An important characteristic of this network design is that any intermediate transformer block generates features of the same dimensions: $\mathbb{R}^{m \times d}$, where $m$ denotes the number of input features or tokens and each lies in $d$ dimensions.



\noindent\textbf{Self-Distilled (SD) ViT for DG:} 
Owing to the monolithic design of ViT it is possible to create several intermediate classifiers. For instance, the output of each transformer block can be provided to the final classifier $h$ to obtain an intermediate classifier: $\mathcal{F}_{j} = f_{j} \circ h$. Whereby each intermediate classifier manifests a feature route through the network. 


Through exploiting the ability to seamlessly create intermediate classifiers, \cite{sultana2022self} developed a technique that randomly samples an intermediate classifier from all the possible ones  at each training iteration. The output of the final classifier is then distilled to this randomly sampled intermediate classifier. 
Specifically, the discrepancy between the final and randomly sampled intermediate classification classifier outputs is computed by comparing the KL divergence between their logit distributions:

\begin{equation} 
\label{eq:kl}
    \mathcal{L}_{\mathrm{KL}}(\mathbf{z}\| \mathbf{z}_j)  =  \sum\limits_{c=1}^C \sigma\left( \mathbf{z}/\tau \right)_c \log\frac{\sigma\left(\mathbf{z}/\tau\right)_{c}}{\sigma\left(\mathbf{z}_j/\tau\right)_{c}}, 
\end{equation} 
where $\mathbf{z}$, $\mathbf{z}_{j} \in \mathbb{R}^{C}$ are the logit vectors produced by $\mathcal{F}$ and $\mathcal{F}_{j}$, respectively, and $C$ is number of classes. $\sigma$ denotes the softmax operation and $\tau$ represent temperature used to rescale the logit vector \cite{hinton2015distilling}. The model is optimized by jointly minimizing the Eq.(\ref{eq:kl}) and $\mathcal{L}_{\mathrm{CE}}$:
\begin{equation}
\label{eq:lossT}
    \mathcal{L} = \mathcal{L}_{\mathrm{CE}} + \lambda \mathcal{L}_{\mathrm{KL}},
\end{equation}
where $\lambda$ balances the contribution of $\mathcal{L}_{\mathrm{KL}}$ towards the overall loss $\mathcal{L}$.






\subsection{Softening Predictions for Self-distillation (SPSD)}
\label{subsection: Preliminaries}

We notice that, in Eq.(\ref{eq:kl}), the $\tau$ is a fixed hyperparameter during training, which is used to rescale the logits from both the full classifier and the sampled intermediate classifier. This can likely act as a bottleneck towards fully harnessing the potential of self-distillation. For instance, it ignores the fact that during self-distillation: (1) the full classifier is not static and it is learning and, (2) during early training, the predictions from both the full and intermediate classifiers are unreliable. It can lead to model overfitting to source domains causing more reliance on the brittle, non-generalizable domain-specific features (Fig.~\ref{fig:motivation_fig}). 
%

\begin{figure*}
\begin{center}
\includegraphics[height=45mm]{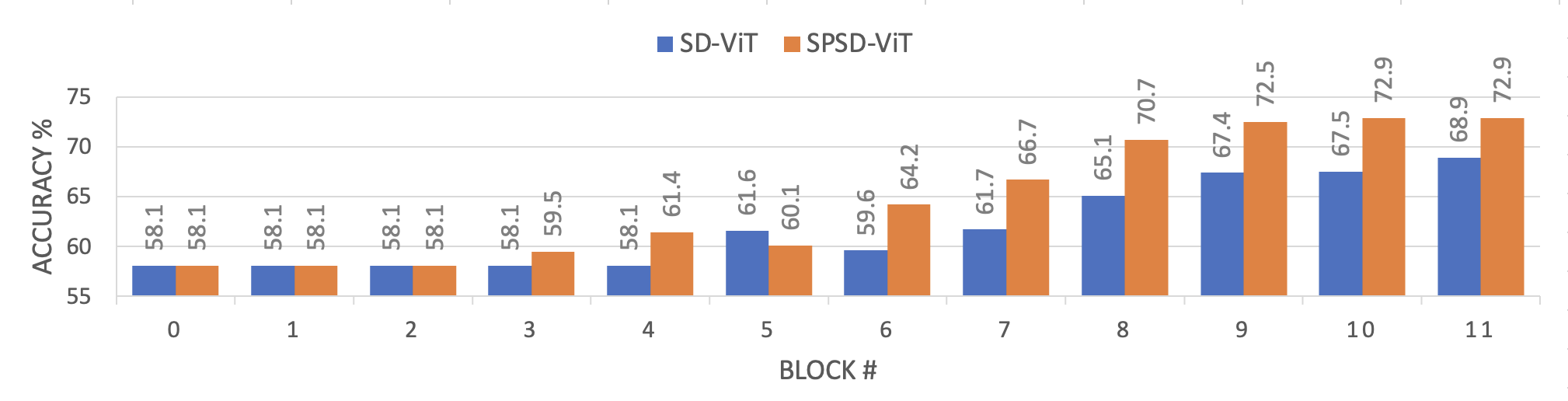}
\end{center}
   \caption{Block-wise accuracy (top-1 \%) for SDViT and ours (SPSD-ViT) on Messidor 2 as target domain.}
\label{fig:intermidiate-block-accuracy} 
\end{figure*}

To this end, inspired by \cite{kim2021self}, we propose to replace this fixed rescaling with an adaptive convex combination of logits vector (from the full classifier or sampled intermediate classifier) with one-hot ground truth vector. As training evolves, this combination gradually increases the influence of the model's own knowledge in the self-distillation process. Fig.~\ref{fig:overall_architecture} displays the overall architecture of our method. We believe that this would better allow enhancing the intermediate feature routes via the process of self-distillation (Fig.~\ref{fig:intermidiate-block-accuracy}), thereby facilitating the learning of cross-domain generalizable features.



\noindent\textbf{Adaptive convex combination:} Let $\mathbf{z}$ be the logit vector produced by the full/intermediate classifier and let $\mathbf{y}$ be the one-hot vector representation corresponding to ground-truth label $y$. We propose an adaptive convex combination of both $\mathbf{z}$ and $\mathbf{y}$ to generate the soft prediction from the full/intermediate classifier as: $\beta \mathbf{z} + (1-\beta) \mathbf{y}$, where $\beta$ is a mixing coefficient that determines how much the model should trust its own prediction from the full network or intermediate classifier. 

\begin{figure*}[!htp]
\begin{center}
\includegraphics[scale=0.48]{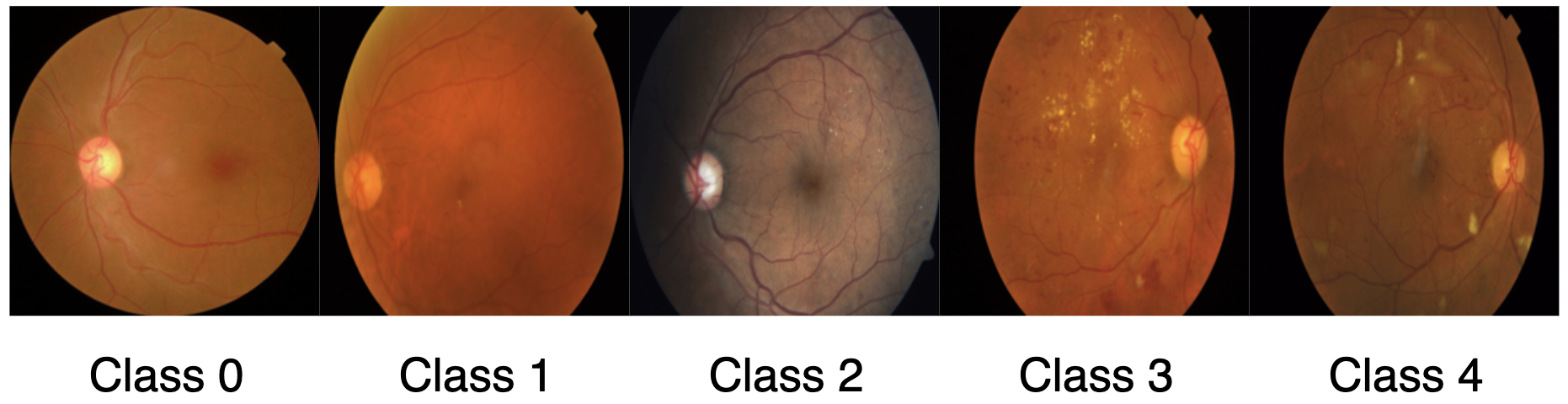}
\end{center}
   \caption{Sample of Images taken from the aptos dataset. Images contain many artefacts such as being out of focus, underexposed, overexposed to light, etc.}
\label{fig:sample}
\end{figure*}

We now discuss how to set the value of $\beta$. The full classifier along with the intermediate classifier does not stay fixed during training and they are constantly evolving. 
Any fixed value of $\beta$ would become sub-optimal at some iteration during training. As discussed earlier, a model is mostly unreliable during the early phases of the training since it has not gone through the data enough times to be able to generate reliable predictions. Therefore, we make $\beta$ as a function of training evolution. The $\beta$ is gradually increased as the training progresses to reflect the increasing reliability of the model. Specifically, the $\beta$ at $t^{th}$ training iteration is computed as: $\beta_{t} =  \beta_{T} \times \frac{t}{T}$ where $T$ is the total number of training iterations and $\beta_{T}$ is the $\beta_{t}$ at final iteration. Note that, $\beta_{T}$ is the only hyperparameter to be sought using the validation set. With $\beta_{t}$, our proposed convex combination becomes adaptive and is now expressed as: $\beta_{t} \mathbf{z} + (1-\beta_{t}) \mathbf{y}$. After including our proposed soft prediction, the KL divergence between the outputs of the full classifier and the randomly sampled intermediate classifier can be computed as:

\begin{equation} 
\label{eq:kl_ours}
   \resizebox{1.0\hsize}{!}{ $\mathnormal{L}_{\mathrm{KL}}(\mathbf{z}\| \mathbf{z}_j)  =  \sum\limits_{c=1}^C \sigma\left( \beta_{t} \mathbf{z} + (1-\beta_{t}) \mathbf{y}\right)_c \log\frac{\sigma\left(\beta_{t} \mathbf{z} + (1-\beta_{t}) \mathbf{y}\right)_{c}}{\sigma\left(\beta_{t} \mathbf{z}_{j} + (1-\beta_{t}) \mathbf{y}\right)_{c}}.$ }
\end{equation}

\begin{table*}[!htp]
  \begin{center}
    {\setlength{\arrayrulewidth}{0.001mm}
 \setlength{\tabcolsep}{8pt}
\renewcommand{\arraystretch}{1}
    {
\begin{tabular}{lcccccc}
\toprule
Method & Backbone\tiny{(\#Param)}  & Aptos & Eyepacs & Messidor & Messidor 2 & Avg.  \\
\hline
ERM\cite{vapnik1999nature}  & ResNet50\tiny{(23.5M)} & 47.6$\pm$1.7 & 71.3$\pm$0.3 & 63.0$\pm$0.4 & 69.0$\pm$1.5  & 62.7 \\
IRM\cite{arjovsky2019invariant}  & ResNet50 & 52.1$\pm$1.7 & 73.2$\pm$0.3 & 51.3$\pm$3.8 & 57.2$\pm$1.7  & 58.4  \\
ARM\cite{zhang2021adaptive}  & ResNet50 & 45.6$\pm$1.5 & 71.7$\pm$0.5 & 62.4$\pm$1.0 & 60.0$\pm$3.4  & 59.9  \\
Fish\cite{shi2021gradient}  & ResNet50 &  44.6$\pm$2.2 & 72.7$\pm$0.7 & 62.1$\pm$0.7 & 66.4$\pm$1.7  & 61.4\\ 
Fishr\cite{rame2021ishr}  & ResNet50 & 47.0$\pm$1.8 & 71.9$\pm$0.6 & 63.3$\pm$0.5 & 66.4$\pm$0.2  & 62.2 \\
GroupDRO\cite{sagawa2019distributionally}  & ResNet50 & 44.9$\pm$3.8 & 72.0$\pm$0.4 & 63.1$\pm$0.9 & 67.8$\pm$1.9  & 62.0 \\
MLDG\cite{Li2018MLDG}  & ResNet50 & 44.1$\pm$1.6 & 71.9$\pm$0.5 & 62.7$\pm$0.6 & 64.4$\pm$0.4  & 60.8\\
Mixup\cite{yan2020improve}  & ResNet50 & 47.3$\pm$1.7 & 72.0$\pm$0.3 & 59.8$\pm$2.8 & 65.8$\pm$1.4  & 61.2\\
Coral\cite{sun2016deep}  & ResNet50 & 44.8$\pm$2.2 & 71.7$\pm$0.9 & 58.6$\pm$2.8 & 68.2$\pm$0.6  & 63.2 \\
MMD\cite{li2018domain}  & ResNet50& 49.3$\pm$0.1 & 69.3$\pm$1.1 & 64.6$\pm$1.4 & 69.6$\pm$0.6  & 60.8 \\
DANN\cite{ganin2016domain}  & ResNet50 & \textbf{54.4$\pm$0.8} & 72.9$\pm$0.4& 57.0$\pm$1.1 & 58.6$\pm$1.7  & 60.7  \\
CDANN\cite{li2018deep}  & ResNet50 & 48.1$\pm$0.7 & 73.1$\pm$0.3& 55.8$\pm$1.8 & 61.2$\pm$1.3  & 59.5  \\
VREX\cite{krueger2021out}  & ResNet50 & 49.6$\pm$2.3 & 73.2$\pm$0.3& 58.5$\pm$0.6 & 65.4$\pm$1.6  & 61.7  \\
SagNet\cite{Nam_2021_CVPR}  & ResNet50 & 41.4$\pm$3.5 & 70.9$\pm$0.9 & 60.8$\pm$0.3 & 66.1$\pm$0.7  & 59.8  \\
RSC \cite{huang2020self} & ResNet50 & 46.7$\pm$0.6 & 71.7$\pm$0.9& 62.5$\pm$0.3 & 66.4$\pm$1.5  & 61.8 \\
SWAD\cite{cha2021swad}  & ResNet50 & 43.8$\pm$2.2 & 71.6$\pm$1.3& 58.9$\pm$1.7 & 67.7$\pm$2.0  & 60.5  \\
DRGen\cite{atwany2022drgen} & ResNet50 & 51.2$\pm$2.1 & 72.6$\pm$0.8 & 59.1$\pm$1.8 & 65.2$\pm$0.6  & 62.1\\
\hline
ERM-ViT\cite{vapnik1999nature}  & DeitSmall\tiny{(22M)} & 48.5$\pm$0.9 & 70.7$\pm$0.7 & 62.7$\pm$1.6 & 69.5$\pm$2.5  & 62.9\\
ERM-ViT\cite{vapnik1999nature}   & T2T-14\tiny{(21.5M)} & \underline{54.0$\pm$3.0} & 73.2$\pm$0.4 & 60.8$\pm$1.7 & 72.0$\pm$0.2  & 62.5\\
ERM-ViT\cite{vapnik1999nature}   & CvT-13\tiny{(20M)}& 49.3$\pm$3.8 & 69.3$\pm$0.1 & \underline{65.2$\pm$0.5} & 70.6$\pm$1.8 & 63.6\\

SD-ViT\cite{sultana2022self}    & DeitSmall\tiny{(22M)} & 48.2$\pm$2.5 & 69.6$\pm$1.5 & 61.5$\pm$0.2 & 69.4$\pm$0.2  & 62.2\\

SD-ViT\cite{sultana2022self}    & T2T-14\tiny{(21.5M)} & 46.5$\pm$0.8 & 71.7$\pm$0.7 & 63.9$\pm$0.9 & 71.4$\pm$0.2  & 63.4\\
SD-ViT\cite{sultana2022self}  & CvT-13\tiny{(20M)}& 47.8$\pm$2.3 & 70.9$\pm$0.8 & 63.9$\pm$1.4 & 72.4$\pm$0.6 & 63.1\\

SPSD-ViT{\footnotesize{(ours)}}  & DeitSmall\tiny{(22M)} & 51.6$\pm$1.1 & 73.3$\pm$0.3 & 64.0$\pm$0.4 & \underline{72.9$\pm$0.1}  & \underline{
65.5}\\

SPSD-ViT{\footnotesize{(ours)}}  & T2T-14\tiny{(21.5M)} & 50.0$\pm$2.8 & \textbf{73.6$\pm$0.3} & \textbf{65.2$\pm$0.3} & \textbf{73.3$\pm$0.2}  & \underline{65.5}\\

SPSD-ViT{\footnotesize{(ours)}} & CvT-13\tiny{(20M)}& 51.7$\pm$1.2 & \underline{73.3$\pm$0.2} & 64.8$\pm$0.5 & 72.4$\pm$0.6 &\textbf{ 65.6}\\

\bottomrule
\end{tabular}
}}
\end{center}
\caption{Multi-Source Domain Generalization Results.}
\label{tab1}
\end{table*}

\section{Experiments}
 \label{section:Experiments}

\noindent\textbf{Datasets:} 
Following the DG method for DR classification \cite{atwany2022drgen}, we evaluate the effectiveness of our proposed method on four existing datasets namely, APTOS\cite{aptos}, EYEPACS\cite{eyepacs}, MESSIDOR and MESSIDOR2\cite{ImageAnalStereol1155}.  It should be noted that each dataset has a high class imbalance (e.g.\ 'No DR' class itself takes up to 74\% of the EYEPACS dataset). Each offers 3662, 88702, 1200, and 1744 fundus images, respectively. Also, each dataset has a set of retina images taken under a variety of imaging conditions belonging to 5 classes: no DR, mild DR, moderate DR, severe DR, and proliferative DR. Images contain many artefacts such as being out of focus, underexposed, overexposed to light, etc. Note that, the images in these datasets are collected under multiple clinical settings using various cameras over an extended period of time which introduces more noise and variations (Fig.~\ref{fig:sample}). We consider each dataset as a separate domain to conduct our experiments. We also report results on WildCamelyon\cite{bandi2018detection} dataset which contains histopathological images of breast cancer metastases in lymph node sections taken over 5 hospitals. Each hospital is considered a domain. There are significant variations in each domain that arise from sources such as differences in patient populations, slide staining, and equipment of image acquisition\cite{veta2016shaimaa, KOMURA201834}. Each domain has two classes, namely tumorous and non-tumorous.

\noindent\textbf{Implementation \& training/testing details:}
We follow the rigorous training and evaluation protocol of DomainBed \cite{Gulrajani2021InSO} to allow a fair comparison among methods. We use the same data augmentations proposed in DomainBed\cite{Gulrajani2021InSO}: crops of random size and aspect ratio,
resizing to 224 × 224 pixels, random horizontal flips, random color jitter, grayscaling the image with
10\% probability, and normalization. We consider each dataset as a domain and use the standard training/validation split of 80\%/20\%. 
We use AdamW\cite{loshchilov2018decoupled} optimizer with its default hyperparameters as in DomainBed\cite{Gulrajani2021InSO} for ERM, with a learning rate of 5e-05 and a batch size of 32. We conduct hyperparameter search only for our model-specific hyperparameters $\lambda$ and $\beta$ in the range of \{0.1 0.3 0.5 0.7 0.9\} and \{0.2 0.4 0.6 0.8\}, respectively. We report classification top-1 accuracy (\%) on each target domain averaged over three different trials with 3 different seeds. Following \cite{Gulrajani2021InSO}, we use the training domain validation settings (IID) as our model selection criteria. Initially, each training domain is split into a subset of training and validation and then we pool all the validation subsets of each training domain to create an overall validation set. The model maximizing the accuracy on the overall validation set is selected as the best model. 
We use PyTorch\cite{https://doi.org/10.48550/arxiv.1912.01703} for implementation and train on 2 V100 GPUs. Table~\ref{tabtrain} shows that our method adds very little training overhead while significantly improving the DG capabilities.

\begin{table}[!h]
  \begin{center}
    {\setlength{\arrayrulewidth}{0.01mm}
 \setlength{\tabcolsep}{4pt}
\renewcommand{\arraystretch}{1}
    \small{
\begin{tabular}{ lc}
\hline

Method  & Average step time  \\
\hline
ERM-ViT \cite{vapnik1999nature}  & 0.353 \\
SD-ViT\cite{sultana2022self}  & 0.361 \\
SPSD-ViT{\footnotesize{(ours)}} & 0.368 \\
\hline
\end{tabular}
}}
\end{center}
\caption{Training overhead in terms of average step time (secs.),}
\label{tabtrain}
\end{table}

\begin{table*}[!htp]
  \begin{center}
    {\setlength{\arrayrulewidth}{0.01mm}
 \setlength{\tabcolsep}{8pt}
\renewcommand{\arraystretch}{1}
    \small{
\begin{tabular}{ lcccccc}
\hline

Method  & Hosp. 0 & Hosp. 1 & Hosp. 2 & Hosp. 3 & Hosp. 4 & Average  \\
\hline
ERM-ViT\cite{vapnik1999nature}  & \underline{97.4$\pm$0.5} & 93.5$\pm$0.8 & 94.5$\pm$0.4 & 96.2$\pm$0.3 & \underline{92.7$\pm$0.9} & 94.9\\
SD-ViT\cite{sultana2022self}  & 97.3$\pm$0.1 & \underline{93.6$\pm$0.5} & \textbf{96.2$\pm$0.6} & \underline{96.3$\pm$0.3} & 91.6$\pm$1.3 & \underline{95.0}\\
SPSD-ViT{\footnotesize{(ours)}} & \textbf{97.4$\pm$0.1} & \textbf{95.3$\pm$0.2} & \underline{95.8$\pm$0.4} & \textbf{96.4$\pm$0.2} & \textbf{93.5$\pm$0.9} & \textbf{95.7}\\

\bottomrule
\end{tabular}
}}
\end{center}
\caption{Multi-Source Domain Generalization Results (IID) on WildCamelyon\cite{bandi2018detection} dataset.}
\label{tabhisto}
\end{table*}


\noindent\textbf{Evaluation with different ViT backbones:}
We conduct experiments with three different ViT-based backbones, namely DeiT\cite{pmlr-v139-touvron21a}, CvT\cite{cvt}, and T2T-ViT\cite{yuan2021tokens} to establish the applicability and generalizability of our method. DeiT is a data-efficient image transformer trained on ImageNet\cite{5206848} using a student-teacher strategy. Note that, we do not utilize the distillation token and the student-teacher formulation settings in our experiments. We use the DeiT-small model having 22M parameters as our default ViT backbone, unless otherwise specified. The DeiT-small model can be regarded as the ViT counterpart of the ResNet-50\cite{DBLP:journals/corr/HeZRS15} which has 23.5M parameters. CvT\cite{cvt} improves vision transformer performance and efficiency by introducing convolutions to ViTs to get the best out of both designs. We consider CvT-13 trained on ImageNet\cite{5206848} for our experiments which has 20M parameters in size. T2T-ViT\cite{yuan2021tokens} propose a method to encode the local structure of the surrounding token and to reduce the length of tokens iteratively by relying on progressive tokenization. The T2T-14 model trained on ImageNet\cite{5206848} containing 21.5M parameters is adopted as the backbone in our experiments. Table~\ref{tab1} shows that our proposed method achieves superior results over all other existing methods irrespective of the backbone architecture used. 

\begin{table}[!h]
  \begin{center}
    {\setlength{\arrayrulewidth}{0.01mm}
 \setlength{\tabcolsep}{4pt}
\renewcommand{\arraystretch}{1}
    \small{
\begin{tabular}{lcccc}
\toprule

Method  & Aptos & Eyepacs & Messidor 2 & Average  \\
\hline

DRGen \cite{atwany2022drgen}  & 41.7$\pm$4.3 & 43.1$\pm$7.9 & 44.8$\pm$0.9  & 43.2\\
ERM-ViT\cite{vapnik1999nature}  & \underline{45.3$\pm$1.3} & 52.4$\pm$3.2 & \underline{58.2$\pm$3.2}  & \underline{51.9}\\
SD-ViT\cite{sultana2022self} & 44.3$\pm$0.9 & \underline{53.2$\pm$1.6} & 57.8$\pm$2.4  & 51.7\\
SPSD-ViT{\footnotesize{(ours)}}   & \textbf{48.3$\pm$1.1} & \textbf{57.4$\pm$2.1} & \textbf{62.2$\pm$1.6}  & \textbf{55.9}\\

\bottomrule
\end{tabular}
}}
\end{center}
\caption{Single-source domain generalization results on Messidor dataset.}
\label{tab2}
\end{table}

\begin{table}[!h]
  \begin{center}
    {\setlength{\arrayrulewidth}{0.01mm}
 \setlength{\tabcolsep}{4pt}
\renewcommand{\arraystretch}{1}
    \small{
\begin{tabular}{lcccc}
\toprule

Method  & Aptos & Eyepacs & Messidor & Average  \\
\hline

DRGen\cite{atwany2022drgen}  &  40.9$\pm$3.9 & \underline{69.3$\pm$1.}0 & \underline{61.3$\pm$0.8 }&  57.7 \\
ERM-ViT \cite{vapnik1999nature} &  47.9$\pm$2.1 & 67.4$\pm$0.9 & 59.6$\pm$3.9  &  58.3\\
SD-ViT  \cite{sultana2022self} &\underline{51.8$\pm$0.9} & 68.7$\pm$0.6 & \textbf{62.0$\pm$1.7} & \underline{60.8}\\
SPSD-ViT{\footnotesize{(ours)}} &  \textbf{52.8$\pm$2.0} & \textbf{72.5$\pm$0.3} & 61.0$\pm$0.8 & \textbf{62.1}\\

\bottomrule
\end{tabular}
}}
\end{center}
\caption{Single-source domain generalization results on Messidor 2 dataset.}
\label{tab22}
\end{table}

\noindent\textbf{Multi-source DG results:}
We report an extensive comparison with the existing SOTA methods in DG literature on DR datasets as shown in Table~\ref{tab1}. We believe that our experiments will offer insights into how the existing SOTA DG methods on natural datasets behave on DR datasets. Moreover, we compare our proposed method with existing SOTA methods\cite{atwany2022drgen} in the DR context. We report ERM results with both CNN and ViT backbones as a baseline as it shows competitive performance against many existing DG methods \cite{Gulrajani2021InSO}. 
We achieve a notable \textbf{+2.1\%} increase in (overall) average accuracy over the second-best contestant. 


\begin{table}[!htp]
  \begin{center}
    {\setlength{\arrayrulewidth}{0.01mm}
 \setlength{\tabcolsep}{4pt}
\renewcommand{\arraystretch}{1}
    \small{
\begin{tabular}{lcccc}
\toprule

Method  & Eyepacs & Messidor & Messidor 2 & Average  \\
\hline

DRGen \cite{atwany2022drgen} &  67.5$\pm$1.8 & \textbf{46.7$\pm$0.1} & \textbf{61.0$\pm$0.1}  & 58.4\\
ERM-ViT\cite{vapnik1999nature}  & 67.8$\pm$1.4 & 45.5$\pm$0.2 & 58.8$\pm$0.4  & \underline{57.3}\\
SD-ViT \cite{sultana2022self} & \textbf{72.0$\pm$0.8} & 45.4$\pm$0.1 & 58.5$\pm$0.2  & \textbf{58.6}\\
SPSD-ViT{\footnotesize{(ours)}} &\underline{71.4$\pm$0.8} & \underline{45.6$\pm$0.1} & \underline{58.8$\pm$0.2 } & \textbf{58.6}\\
\bottomrule
\end{tabular}
}}
\end{center}
\caption{Single-source domain generalization results on Aptos dataset.}
\label{tabaptos}
\end{table}

\begin{table}[!htp]
  \begin{center}
    {\setlength{\arrayrulewidth}{0.01mm}
 \setlength{\tabcolsep}{4pt}
\renewcommand{\arraystretch}{1}
    \small{
\begin{tabular}{lcccc}
\toprule

Method  & Aptos & Messidor & Messidor 2 & Average  \\
\hline

DRGen \cite{atwany2022drgen} & 61.3$\pm$1.9 &  \textbf{54.6$\pm$1.5} & \textbf{65.4$\pm$0.1}  & 60.4\\
ERM-ViT \cite{vapnik1999nature} & 69.1$\pm$1.4 &  50.4$\pm$0.3 & 62.8$\pm$0.2  & \underline{60.8}\\
SD-ViT\cite{sultana2022self}  &  \underline{69.3$\pm$0.3} & 50.0$\pm$0.5 & \underline{62.9$\pm$0.2}  & 60.7\\
SPSD-ViT{\footnotesize{(ours)}} & \textbf{75.1$\pm$0.5} & \underline{50.5$\pm$0.8} & 62.2$\pm$0.4  & \textbf{62.5}\\
\bottomrule
\end{tabular}
}}
\end{center}
\caption{Single-source domain generalization results on Eyepacs.} dataset.
\label{tabeye}
\end{table}

\noindent\textbf{Single-source DG results:}
In the single-source DG setting, we train our model only on one dataset and evaluate on the other 3 datasets. The results in Table~\ref{tab2},\ref{tab22},\ref{tabaptos},\ref{tabeye} show that our method achieves superior performance in all the datasets even under heavy class-imbalance. Single-source DG is even more challenging in messidor dataset as the model does not have any data for class id 4 (proliferative DR). Table~\ref{tab2} shows that our method can generalize to new unseen domains even under a heavy class-imbalanced scenario with a significant improvement in the performance (+4.2\%). 

\noindent\textbf{DG capability in other medical images:}
We also show results for multi-source domain generalization results on the WildCamelyon dataset \cite{bandi2018detection}. This dataset constitutes an extensive collection of histopathological images representing breast cancer metastases. The detailed results can be found in Table~\ref{tabhisto}.
Our method SPSD-ViT achieves superior results with 95.7\% accuracy over both ERM-ViT and SD-ViT. These results underscore the versatility of our proposed SPSD-ViT method. While the core scope of our study includes only the Diabetic Retinopathy (DR) data, we found that the method's application is not limited to this area. 

%

\noindent\textbf{Calibration performance:}
We also evaluate the calibration performances of our proposed method under multi-source settings, utilizing two widely accepted evaluation metrics to measure the miscalibration of a model - Expected Calibration Error (ECE)\cite{conf/cvpr/NixonDZJT19}, and Static Calibration Error (SCE)\cite{Naeini2015ObtainingWC}. The calibration performance of a predictive model is crucial in decision-making processes, which are an important part of healthcare applications. It establishes the credibility and reliability of the model's predictions.
Table~\ref{tabcal} illustrates the superior performance of our method compared to established baselines. Notably, our method not only exceeds the baselines in domain generalization (DG) performance, but also shows superior calibration performance. This dual achievement underscores the method's robustness, reinforcing the reliability of its predictions while maintaining high performance levels. This makes our method a promising tool for effective, reliable decision-making in various applications e.g., healthcare.

\begin{table*}[!htp]
  \begin{center}
  \scalebox{0.95}{
    {\setlength{\arrayrulewidth}{0.01mm}
 \setlength{\tabcolsep}{2pt}
\renewcommand{\arraystretch}{1}
    \small{
\begin{tabular}{@{}lcccccccc@{}}

\toprule
\multicolumn{1}{c}{\multirow{2}{*}{\textbf{Dataset}}} & 
\multicolumn{2}{c}{Aptos}  & 
\multicolumn{2}{c}{Eyepacs}  &
\multicolumn{2}{c}{Messidor}  &
\multicolumn{2}{c}{Messidor 2}\\ 
\cmidrule{2-9}\\[-1em]

  & ECE & SCE & ECE & SCE & ECE & SCE & ECE & SCE\\
\midrule
ERM-ViT\cite{vapnik1999nature}       & $33.80\pm2.60$ & $16.21\pm1.40$  & $22.66\pm1.18$ & $9.88\pm0.34$ & $25.40\pm3.69$ & $10.94\pm1.11$ & $16.88\pm2.74$ & $8.19\pm1.02$\\
SD-ViT\cite{sultana2022self}       & $\underline{27.13\pm1.03}$ & $\underline{15.04\pm2.20}$ & $\mathbf{16.97\pm1.77}$ & $\underline{8.31\pm0.14}$ & $\underline{22.22\pm1.73}$ & $\underline{10.12\pm0.83}$ & $\underline{12.67\pm1.92}$ & $\underline{7.08\pm0.40}$\\
SPSD-ViT{\footnotesize{(ours)}} & $\mathbf{23.20\pm4.00}$ & $\mathbf{14.11\pm1.41}$ & $\underline{17.06\pm2.12}$ & $\mathbf{7.88\pm0.99}$ & $\mathbf{20.86\pm1.84}$ & $\mathbf{9.41\pm0.47}$ & $\mathbf{11.36\pm2.77}$ & $\mathbf{6.41\pm0.47}$\\

\bottomrule
\end{tabular}
}}}
\end{center}
\caption{Calibration performance in SCE\cite{Naeini2015ObtainingWC} and ECE\cite{conf/cvpr/NixonDZJT19} (in scale of $10^{-2}$).}
\label{tabcal}
\end{table*}

\begin{table*}[!htp]
  \begin{center}
    {\setlength{\arrayrulewidth}{0.01mm}
 \setlength{\tabcolsep}{8pt}
\renewcommand{\arraystretch}{1}
    \small{
\begin{tabular}{lccccc}
\hline
Method  & Aptos & Eyepacs & Messidor & Messidor 2 & Average  \\
\midrule
None & 48.2$\pm$2.5 & 69.6$\pm$1.5 & 61.5$\pm$0.2 & 69.4$\pm$0.2  & 62.2 \\
Final classifier only &  \underline{51.5$\pm$0.4} & \textbf{73.5$\pm$0.1} & 60.7$\pm$0.7 & \underline{69.7$\pm$1.7 } & \underline{63.9}\\
Interm. classifier only  &  49.1$\pm$0.4 & 71.7$\pm$0.5 & \underline{63.0$\pm$2.4} & 68.4$\pm$3.4  & 63.0\\ 
Final \& interm. classifier (SPSD-ViT) & \textbf{51.6$\pm$1.1} & \underline{73.2$\pm$0.3} & \textbf{64.0$\pm$0.4} & \textbf{72.9$\pm$0.1}  & \textbf{65.5}\\
\bottomrule
\end{tabular}
}}
\end{center}
\caption{Ablation studies on proposed prediction softening.}
\label{tabsoft}
\end{table*}

\begin{table*}[!htp]
  \begin{center}
    {\setlength{\arrayrulewidth}{0.01mm}
 \setlength{\tabcolsep}{8pt}
\renewcommand{\arraystretch}{1}
    \small{
\begin{tabular}{ lccccc  }
\hline

Hyper-parameters & Aptos & Eyepacs & Messidor & Messidor 2 & Average  \\
\midrule
$\lambda$ = 0.1 & 47.7$\pm$0.7 & 70.9$\pm$1.2 & 63.3$\pm$0.2 & 71.9$\pm$0.7  & 63.5 \\
$\lambda$ = 0.3 & 50.6$\pm$1.2 & 73.2$\pm$0.3 & 60.2$\pm$2.0 & 71.0$\pm$0.7  & 63.7 \\
$\lambda$ = 0.5 & \textbf{52.7$\pm$0.8} & \textbf{73.5$\pm$0.2} & 61.2$\pm$2.8 & 71.9$\pm$0.3  & 64.8 \\
$\lambda$ = 0.7\tiny{(Default)}  & \underline{ 51.6$\pm$1.1} & 73.3$\pm$0.7 &\underline{ 64.0$\pm$0.4 }& \textbf{72.9$\pm$0.1}  & \textbf{65.5}\\ 
$\lambda$ = 0.9 & 50.5$\pm$0.7 &\underline{ 73.4$\pm$0.4} & \textbf{64.5$\pm$0.3} & \underline{72.2$\pm$0.5}  &\underline{ 65.2 }\\
\midrule
Fixed $\beta$ = 0.5  &  44.2$\pm$1.9 & 71.6$\pm$1.0 & 62.0$\pm$1.4 & 71.3$\pm$0.3  & 62.2\\ 
$\beta$ = 0.2  &  48.2$\pm$0.4 & 69.6$\pm$0.9 &\underline{ 63.9$\pm$0.2 }& \underline{72.6$\pm$0.6}  & 63.5\\ 
$\beta$ = 0.4  &  47.3$\pm$1.6 & 68.1$\pm$0.7 & 62.9$\pm$1.2 & 71.7$\pm$0.6  & 62.5\\ 
$\beta$ = 0.6   &  \textbf{53.1$\pm$0.6} &72.5$\pm$0.3 & 63.4$\pm$0.9 & 71.9$\pm$0.8  & \underline{65.2}\\
$\beta$ = 0.8\tiny{(Default)}&  \underline{51.6$\pm$1.1} & \textbf{73.3$\pm$0.7} & \textbf{64.0$\pm$0.4} & \textbf{72.9$\pm$0.1}  & \textbf{65.5}\\ 
$\beta$ = 0.1 & 49.7$\pm$0.8       &   \underline{ 73.2$\pm$0.2 }   &      63.5$\pm$0.6      &    72.3$\pm$0.6     &     64.7 \\
\bottomrule
\end{tabular}
}}
\end{center}
\caption{Detailed results on the sensitivity of SPSD-ViT(ours) to $\lambda$ and $\beta$.}
\label{tabhype}
\end{table*}



\noindent\textbf{Ablations on prediction softening and hyperparameter analysis:}
We show results with possible ablations of our prediction softening in Table~\ref{tabsoft}. In the final classifier only case, logits are softened at the final block only using our adaptive convex combination, and in the intermediate classifier only case, logits are softened at the randomly selected intermediate block. Results show that our prediction softening on both full and intermediate classifier, as proposed, achieves superior results compared to applying only on either the full or the intermediate classifier. 
In Table~\ref{tabhype}, we present a detailed analysis focusing on the sensitivity of two key parameters in our method: $\lambda$ derived from Equation (\ref{eq:lossT}), and $\beta$ which features in our adaptive convex combination. 
We note that, our method is relatively resilient to minor perturbations in the $\beta$ and $\lambda$ parameters, continuing to deliver comparable performance even when these variables deviate slightly from the best-found values. This highlights the stability of our method and suggests that it is not overly reliant on hyperparameter fine-tuning. However, it is notable that constraining the value of $\beta$ to a fixed level tends to lead to sub-optimal performance, emphasizing the importance of this adaptive parameter.
Note that, all these experiments are conducted using the DeiT-small backbone. 

\section{Conclusion}
\label{section:Conclusion}
We present a new DG approach for DR classification based on distilling the model's own knowledge to its intermediate blocks by constructing a new prediction softening scheme, which is an adaptive convex combination of one-hot labels and the model's own knowledge. 
We reported comprehensive results derived from multiple Diabetic Retinopathy (DR) datasets, with both multi-source and single-source domain generalization (DG) settings, in conjunction with various Vision Transformer (ViT) backbones. These wide-ranging experiments corroborate the effectiveness and versatility of our method against previously established techniques.
Beyond delivering superior DG performance (top-1 accuracy), our method also shows improved out-domain calibration performance (ECE and SCE).
Notably, the importance of these improved calibration performances cannot be overstated in a number of safety-critical applications, with healthcare being a prime example. In such critical fields, the reliability of a model's prediction is of paramount importance. 


%

{\small
\bibliographystyle{ieee_fullname}
\bibliography{egbib}
}

\end{document}